\pdfoutput=1

\documentclass[11pt]{article}

\usepackage[]{acl}

\usepackage{times}
\usepackage{latexsym}

\usepackage[T1]{fontenc}

\usepackage[utf8]{inputenc}

\usepackage{microtype}
\usepackage{graphicx}
\usepackage{multirow}
\usepackage{subfigure}
\title{GEB-1.3B: Open Lightweight Large Language Model}

\author{Jie Wu, Yufeng Zhu, Lei Shen, Xuqing Lu \\ 
  GEB\\
}

\begin{document}
\maketitle

\begin{abstract}
Recently developed large language models (LLMs) such as ChatGPT, Claude, and Llama have demonstrated impressive abilities, and even surpass human-level performance in several tasks. Despite their success, the resource-intensive demands of these models, requiring significant computational power for both training and inference, limit their deployment to high-performance servers. Additionally, the extensive calculation requirements of the models often lead to increased latency in response times. With the increasing need for LLMs to operate efficiently on CPUs, research about lightweight models that are optimized for CPU inference has emerged.
In this work, we introduce GEB-1.3B, a lightweight LLM trained on 550 billion tokens in both Chinese and English languages. We employ novel training techniques, including ROPE, Group-Query-Attention, and FlashAttention-2, to accelerate training while maintaining model performance. Additionally, we fine-tune the model using 10 million samples of instruction data to enhance alignment.
GEB-1.3B exhibits outstanding performance on general benchmarks such as MMLU, C-Eval, and CMMLU, outperforming comparative models such as MindLLM-1.3B and TinyLLaMA-1.1B. Notably, the FP32 version of GEB-1.3B achieves commendable inference times on CPUs, with ongoing efforts to further enhance speed through advanced quantization techniques. The release of GEB-1.3B as an open-source model marks a significant contribution to the development of lightweight LLMs, promising to foster further research and innovation in the field.
\end{abstract}

\section{Introduction}

Large language models have experienced significant advancements, achieving superhuman capabilities in numerous specific tasks. These models, such as ChatGPT 3.5~\citep{openai_chatgpt}, GPT-4~\citep{openai2024gpt4}, and Claude~\citep{claude}, demonstrate remarkable success across various languages, particularly in English. On the other hand, models such as Baichuan~\citep{baichuan2023baichuan2}, ChatGLM~\citep{zeng2022glm}, and Qwen~\citep{qwen} are mainly used for Chinese. However, these powerful models require a lot of computer resources, like GPUs or TPUs, for both training and inference phases. Consequently, these models are typically operational on remote high-performance servers. Furthermore, the vast computations required by the models bring potential delays in responding. There exists a significant demand for models that can efficiently run on CPUs, thereby enabling deployment on more accessible devices such as laptops and smartphones. In response to this need, researchers are now focusing on the development of lightweight large language models, aiming to reduce both response times and hardware costs. Remarkably, the Llama 2 model~\citep{Touvron2023Llama2O} series demonstrates that, when trained with massive data, smaller models can surpass their larger counterparts in performance. Despite being trained on trillions of tokens, the Llama 2 series shows no signs of training loss saturation, suggesting untapped potential in smaller model capacities.

In this work, we introduce GEB-1.3B, a lightweight model featuring 1.3 billion parameters and trained on 550 billion tokens in both Chinese and English languages. We incorporate advanced techniques such as ROPE~\citep{su2024roformer}, Group-Query-Attention~\citep{ainslie2023gqa}, and FlashAttention-2~\citep{dao2022flashattention} to expedite our training process. Additionally, we fine-tune the model using 10 million instruction-based samples to enhance its alignment.
\begin{table*}
\centering
\begin{tabular}{lccccc}
\hline
\textbf{Language} & \textbf{Dataset}\ & \textbf{Proportion} & \textbf{Size} &\textbf{Tokens}\\
\hline
\multirow{2}{*}{English}&C4&33.1\%&	493G	&182B\\
&Github+StackExchange	&14.7\%&	245G&	81B\\
\hline
\multirow{3}{*}{Chinese}&CommonCrawl&	28.4\%&	700G&	156B\\
&WuDaoCorpus	&8.9\%&	200G&	49B\\
&SkyPile	&14.9\%&	350G&	82B\\
\hline
\end{tabular}

\caption{The details of the datasets.}
\label{tab:data}
\end{table*}

The evaluation results indicate that GEB-1.3B outperforms comparable models such as MindLLM-1.3B\citep{yang2023mindllm} and TinyLLaMA-1.1B\citep{zhang2024tinyllama} on general benchmarks, including MMLU\citep{hendrycks2020measuring}, C-Eval\citep{huang2024c}, and CMMLU\citep{li2023cmmlu}. Moreover, the inference time of the FP32 version on CPUs is sufficiently rapid for practical applications. And we plan to explore further acceleration through quantization in the future.

We are releasing GEB-1.3B to the general public for research use at \url{https://huggingface.co/GEB-AGI/geb-1.3b}. The model has demonstrated superior performance than other lightweight models of the similar parameter size, and even surpass some larger counterparts. We believe that the release of this lightweight LLM will benefit further research.

The remainder of this paper describes the pre-training methodology (Section \ref{pretrain}), alignment technology (Section \ref{align}), evaluations on benchmarks (Section \ref{eval}), and conclusions (Section \ref{con}). 
\section{Pre-training}
\label{pretrain}
\subsection{Data}

To ensure the coverage, diversity, and quality of foundational datasets for our 1.3 billion parameter model, we employ multifaceted approaches in data collection and processing.
\subsubsection{Data Collection}

\noindent\textbf{English Pre-training Dataset} Our English pre-training dataset predominantly originates from C4~\citep{raffel2020exploring}, which contains over 15.6 billion tokens from 365 million internet domains. To further augment the model's logical capability, we integrate datasets from GitHub and Stack Exchange during the pre-training phase, including code snippets, question-and-answer exchanges, and procedural instructions.

\noindent\textbf{Chinese Pre-training Dataset}
In contrast to the English counterpart, there are few high-quality, large-scale Chinese corpora suitable for pre-training. To bridge this gap, we construct a comprehensive large-scale Chinese pre-training corpus from the Common Crawl dataset, meticulously ensuring both quality and diversity. This corpus adheres to the processing principles applied to the C4 dataset, with additional, tailored cleaning protocols for the Chinese language, as elaborated in Subsection \ref{dataproc}. Moreover, we incorporate two publicly accessible high-quality Chinese corpora - WuDaoCorpus (200G)~\citep{yuan2021wudaocorpora} and SkyPile~\citep{wei2023skywork} - as ancillary data sources, thereby enriching our foundational dataset. Table \ref{tab:data} shows the details of the datasets.
\subsubsection{Data Processing}
\label{dataproc}
The  Common Crawl dataset is crawled from the web and often suffers from quality issues, because of including garbled text, incomplete URL links, extraneous emojis, fragmented sentences, personal contact details, sensitive expressions, and profanity. To obtain informative and diverse data, we undertake multiple stages of processing.

\noindent\textbf{Rule Cleaning}
First, we extract a Chinese corpus from the Common Crawl dataset, by applying specialized rules to remove non-essential HTML, CSS, and JavaScript identifiers, garbled text, and other aberrant symbols. We eliminate incomplete sentences by identifying terminal punctuation and applying a sentence length threshold. Regular expressions are used to remove personal information and URL links, while a word list screens out sentences with sensitive language, profanity, or promotional content.
\begin{table*}
\centering
\begin{tabular}{lccccccc}
\hline
\textbf{Size} & \textbf{Vocabulary Size}\ & \textbf{Hidden Size} & \textbf{FFN Size} &\textbf{Heads} &\textbf{Layers}	&\textbf{KV Groups}&\textbf{Length}\\
\hline
1.3B&	64896&	2048&	5632&	16&	24&	4&	4096\\
\hline
\end{tabular}

\caption{The details of GEB-1.3B.}
\label{tab:model}
\end{table*}

\noindent\textbf{High-Quality Corpus Filtering}
We leverage perplexity (PPL) and keyword density filtering techniques to get rid of low-quality content and advertising. Given that perplexity inversely correlates with sentence fluency, we establish a perplexity threshold to exclude samples failing to meet our quality standards. Additionally, we employ a keyword density-based filtering method to remove sentences with insufficient knowledge density, setting a predetermined threshold to identify and discard the samples below the threshold. Recognizing that internet-sourced corpora often consist of paragraphs with a single sentence, we utilize a classification model to concatenate highly correlated sentences into cohesive paragraphs.

\noindent\textbf{Deduplication}
Our deduplication process involves removing instances of consecutive duplicate sentences and paragraphs in each dataset. Further comparisons across different datasets help eliminate redundant paragraphs, ensuring the uniqueness of the information within our corpus. Through these strict rules and filtering methods, we have compiled a 1.3TB pre-training dataset, which includes 700GB of high-quality Chinese data extracted from the 50TB Common Crawl dataset.

\subsection{Architecture}
The architecture of GEB-1.3B is grounded in the Transformers framework, yet it incorporates several advanced techniques to improve the performance. Herein, we show the details of these adaptations:

\noindent\textbf{Tokenization} 
There are numerous vocabularies available, eliminating the need to retrain the tokenizer model. We decide to exclude the original Llama tokenizer model due to its suboptimal tokenization efficiency for Chinese tasks. By building upon the ChatGLM-3 tokenizer model, we develop a new vocabulary containing 64,896 entries. This is different from the vocabularies of Qwen and Baichuan2, which both exceed 100,000 entries and are deemed too large. The rationale behind this strategy is to utilize a more compact vocabulary, thereby reducing the number of model parameters. This approach aims to enhance the model's inference speed without sacrificing its performance.

\noindent\textbf{Word Embedding Layer} 
Through experiments, we employ an untied embedding strategy, diverging from the conventional practice of tying the weights of the input embedding to the output layer. This choice, though increasing memory requirements, is justified by the resultant performance gains. The Rotary Positional Embedding (RoPE) method \cite{su2024roformer}, serves as our primary mechanism for incorporating positional information into the model. This approach is widely used in modern large-scale language models. Furthermore, we utilize FP32 for generating position IDs, a measure designed to prevent position collisions that may arise from inadequate precision.

\noindent\textbf{Transformer Blocks} 
Our modifications to the transformer blocks are aimed at optimizing training efficiency and stability, particularly within the constraints of limited GPU resources. We choose the Group-Query-Attention (GQA)~\citep{ainslie2023gqa} mechanism rather than the conventional multi-head attention (MHA)~\citep{vaswani2017attention} mechanism. This innovation significantly enhances inference speed without detracting from accuracy, by grouping the heads of K and V to share a singular Q. The SwiGLU~\citep{shazeer2020glu} activation function, a hybrid of Swish~\citep{ramachandran2017searching} and Gated Linear Unit~\citep{dauphin2017language}, is selected for its efficacy. The dimension of the Feed-Forward Networks (FFN) is adjusted from four times the hidden sizes to a ratio of 8/3. Post-RMSNorm~\citep{ba2016layer} is implemented instead of traditional normalization methods, providing stronger regularization and facilitating improved convergence during the training phase. Moreover, all biases within the transformer blocks are eliminated.

The details of GEB-1.3B are depicted in Table \ref{tab:model}.

\subsection{Infrastructure}
\begin{figure}
\centering
\subfigure{
\includegraphics[scale=0.33]{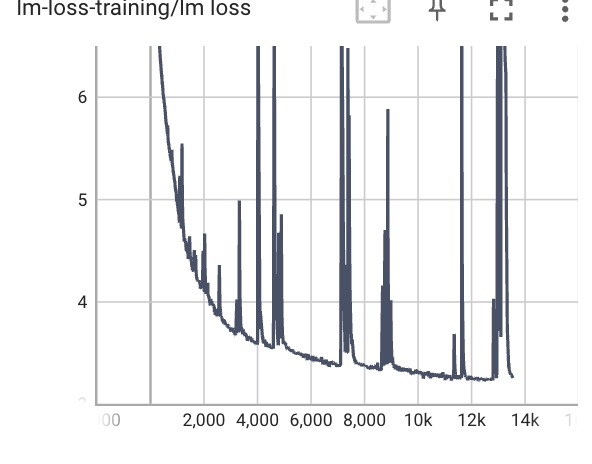}}
\subfigure{
\includegraphics[scale=0.43]{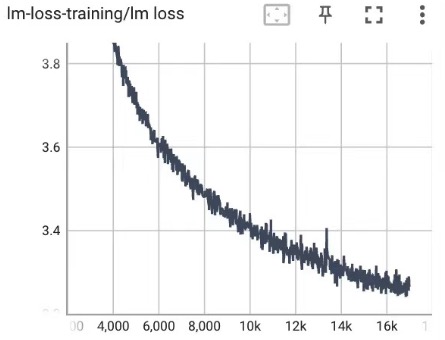}}
\caption{Loss curves before and after adopting four measures}
\label{fig:loss}
\end{figure}

In the pre-training phase, we employ the AdamW~\citep{loshchilov2017decoupled} optimizer, setting the hyperparameters to $\beta_1$ = 0.9, $\beta_2$ = 0.95, and $\epsilon$ = 10e\textendash8. The learning rate follows a cosine decay schedule, with a minimum at 4e-5 and a peak at 4e-4. To ensure stability throughout the training, we utilize BFloat16 mixed precision method. The computational resources are limited to 64 NVIDIA RTX3090ti GPUs. Due to CUDA memory constraints, we establish a global batch size of 320 samples, accommodating a model sequence length of 4096 tokens. Such a small batch size makes training unstable, leading to frequent loss spikes during the pre-training stage. To mitigate this, we implement following strategies and the loss curves are shown in Figure \ref{fig:loss}:

\textbf{Batch Sample Replacement}: We observe that a diverse sample distribution within a batch may lead to a spike in loss. To address this, we replace the affected batch with an alternative batch when a spike occurs.

\textbf{Skipping Specific Iterations}: Drawing inspiration from the practices of PaLM~\citep{chowdhery2023palm} and OPT~\citep{zhang2022opt} models, we exclude samples from the 100-200 iterations preceding and succeeding a loss spike.

\textbf{Embedding Layer Gradient Shrink (EGS)}: Building on the theoretical insights provided by Molybog~\citep{molybog2023theory}, we identify that the emergence of a loss spike is associated with sudden increases in the gradient updates of the shallow layers. This surge in gradients can form a chain reaction, affecting the update status of parameters in the deeper layers of the model and leading it into a non-stationary state~\citep{zeng2022glm}. So we directly multiply the shallow gradient by the scaling coefficient $\alpha$ to reduce the update value of the shallow gradient.
\begin{figure}
    \centering
     \includegraphics[scale=0.33]{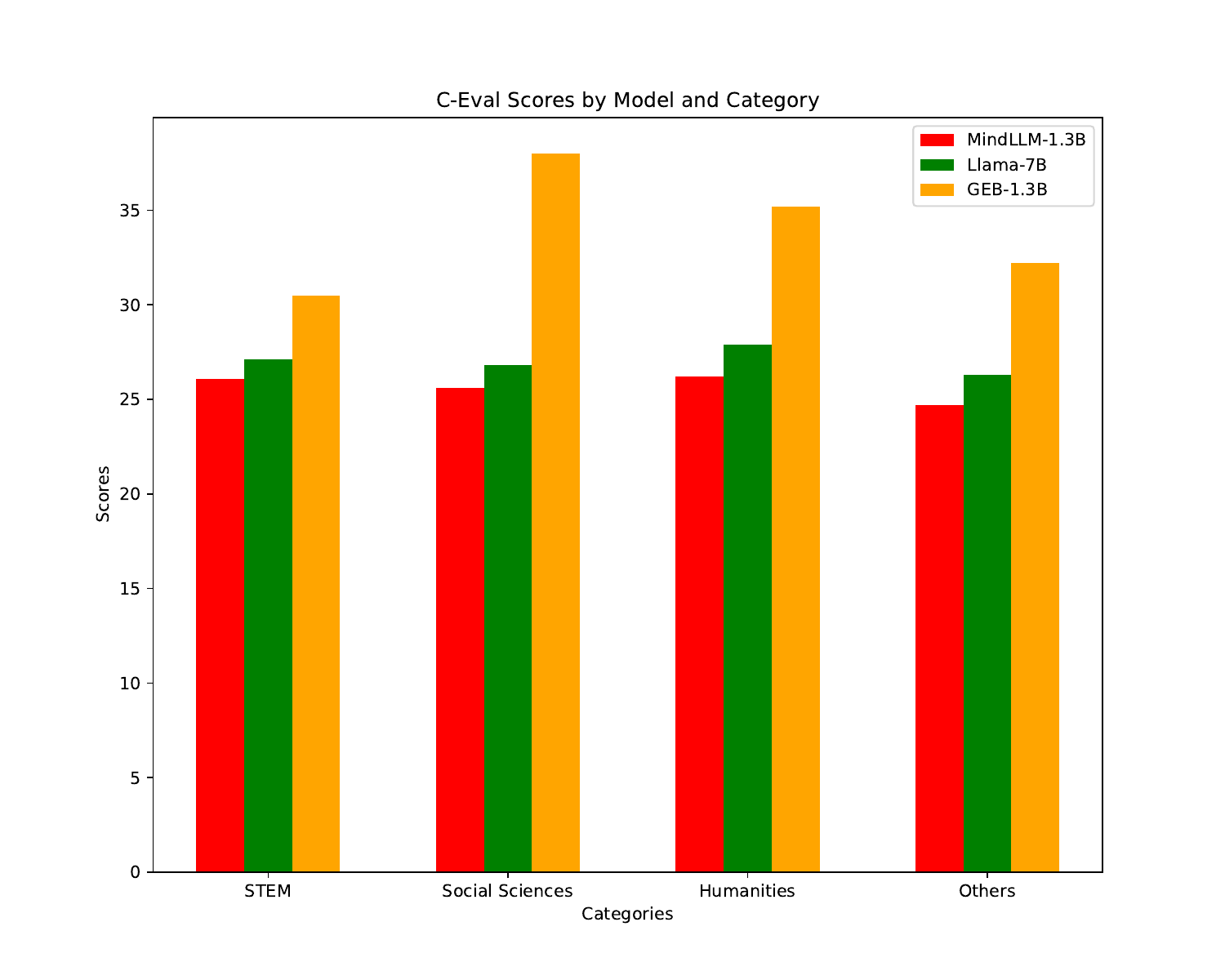}

    \caption{The results on C-Eval benchmark.}
    \label{fig:ceval}
\end{figure}

\textbf{Adjusting Learning Rate}: When reducing the batch size, it is imperative to correspondingly decrease the learning rate to prevent oscillations. Conversely, as the batch size increases, adjusting the learning rate upwards is necessary to expedite convergence. With our global batch size setting at only 320, reducing the learning rate accordingly is essential to avoid loss spikes. Moreover, we have observed that increasing the decay rate of the learning rate can effectively prevent loss spikes.

\section{Alignment}
\label{align}
The pretrained large language models only learn the knowledge of the world, but there is still a gap between the model generations and human expectations. The most recent advances of alignment are Supervised Fine-Tuning (SFT)~\citep{ouyang2022training} and Direct Preference Optimization (DPO)~\citep{rafailov2024direct}. By adopting Supervised Fine-Tuning (SFT) and Direct Preference Optimization (DPO), the GEB-1.3B model is trained to align more closely with human conversational modes.

\begin{table*}
\centering
\begin{tabular}{lcccc}
\hline
\textbf{Model} & \textbf{MMLU}\ & \textbf{C-Eval} & \textbf{CMMLU} &\textbf{Average}\\
\hline
Baichuan-7B&	{42.30}&	{42.80}&	{44.02}&	{43.04} \\
ChatGLM-6B&	{40.63}&	{38.90}&	{-}&	{39.77} \\ 
\textbf{GEB-1.3B}&	{31.20}&	{33.30}&	{32.20}&	{32.23} \\ 
Llama-7B&	{35.10}&	{27.10}&	{26.75}&	{29.65} \\
Falcon-7B&	{	28.00}&	{	-}&	{	-}&	{	28.00}  \\ 
MPT-7B&	{27.93}&	{27.15}&	{26.00}&	{27.03}  \\
MindLLM-1.3B&	{	26.20}&	{	26.10}&	{	25.33}&	{	25.88}  \\ 
MindLLM-3B&	{26.20}&	{25.70}&	{25.00}&	{25.63}  \\ 
TinyLlama-1.1B&	{25.34}&	{25.02}&	{24.03}&	{24.80}  \\ 

\hline
\end{tabular}

\caption{The results on MMLU, C-Eval, and CMMLU benchmarks.}
\label{tab:res}
\end{table*}
\subsection{Supervised Fine-Tuning}
Our SFT dataset comprises approximately 16 million instances of instructional data, spanning a wide list of subjects that range from benign and useful topics to a sensitive one. This collection includes an expansive variety of domains, such as general linguistic tasks, mathematical problem-solving, and programming exercises. On the other hand, the safety-related portion of our dataset encompasses a broad spectrum of sensitive topics. In addition, we craft a series of prompts designed for various tasks, aiming to enhance the model's capability for generalization. This approach ensures both the diversity and coverage of the dataset.
\subsection{Direct Preference Optimization}
We use DPO to align the model with human expectations, in order to guide the model to avoid answer unethical questions and generate harmless answers. The dataset size is only 10k for DPO, which is rather small comparing to the SFT dataset.

\section{Evaluations}
\label{eval}
\subsection{General Benchmarks}
In our evaluation, we compare our model with several well-known large language models, including Llama2-7B, Baichuan-7B, Falcon-7B, MPT-7B, and ChatGLM-6B. Additionally, we extend our comparison to the models such as MindLLM-1.3B~\citep{yang2023mindllm} and TinyLLaMA-1.1B~\citep{zhang2024tinyllama}, whose parameter size is comparable to our own. This multifaceted comparison enables a comprehensive understanding of our model's performance.

We evaluate the models on three general benchmarks, MMLU~\citep{hendrycks2020measuring}, C-Eval~\citep{huang2024c}, and CMMLU~\citep{li2023cmmlu}. The MMLU benchmark, which evaluates the English knowledge, serves as a key metric for assessing the models' performance in English. Conversely, the C-Eval and CMMLU benchmarks focus on Chinese language proficiency. 

The results on the three benchmarks are shown in table \ref{tab:res}.
As table \ref{tab:res} reveals, GEB-1.3B significantly outperforms the LLaMA-7B model. Compared to models of similar scale, such as MindLLM-1.3B and TinyLLaMA-1.1B, the GEB-1.3B model is notably superior, demonstrating exceptional capabilities. In contrast to the LLaMA-7B model, the GEB-1.3B model shows significantly better performance in Chinese, while its English language performance, though slightly lower, is still remarkable. Overall, the comparison not only highlights the GEB-1.3B model's strong performance in both languages but also underscores its superiority over other lightweight models.

Figure \ref{fig:ceval} displays GEB-1.3B, MindLLM-1.3B, and Llama-7B's performances on the C-Eval dataset. GEB-1.3B utilizes a zero-shot Chain of Thought method for answer deduction. GEB-1.3B outperforms MindLLM-1.3B across all fields, and even surpasses Llama-7B which is much larger.
\begin{table}
\centering
\begin{tabular}{lc}
\hline
\textbf{Model} & \textbf{ToxiGen}\\
\hline
GEB-1.3B	&3.00 \\
Falcon-7B	&7.89\\
MPT-instruct&	16.33\\
Llama2-7B	&21.25\\

\hline
\end{tabular}

\caption{The results on the ToxiGen dataset.}
\label{tab:tox}
\end{table}
\subsection{Toxicity}
We also experiment on the ToxiGen~\citep{hartvigsen2022toxigen} dataset, to evaluate the toxicity in the model-generated text. The evaluation covers a variety of models, including GEB-1.3B, Falcon-7B~\citep{almazrouei2023falcon}, MPT-instruct~\citep{MosaicML2023Introducing}, and Llama2-7B. Lower scores reflect better performance. As depicted in table \ref{tab:tox}, despite the small size of the model parameters, GEB-1.3B generates a minimal amount of toxic text. There's no simple correlation between the model size and its effectiveness on the ToxiGen benchmark.

\subsection{Inference Speed}
We assess the inference speed of the model in the CPU environment. The current CPU inference time of the FP32 version reaches 12 tokens/second, which means it could be used in edge devices. We plan to quantify the model in the future for further acceleration.

\section{Conclusion}
\label{con}
In this study, we introduce GEB-1.3B, a novel open-source, lightweight large language model. We detail the methodologies employed in the training and alignment, which are all cutting-edge technologies of large language models. Our experimental findings demonstrate that GEB-1.3B surpasses its counterparts, i.e., the models with similar parameter sizes, on general benchmarks. Furthermore, our results affirm that smaller models, when trained with extensive datasets, can achieve performance comparable to that of larger models. Notably, GEB-1.3B is designed to operate efficiently on CPUs, thereby extending its utility to laptops, smartphones, and other edge devices. This advancement marks a significant stride towards the broader application and accessibility of large language models in various computing environments.

\section{Limitations and Ethical Considerations}

Our model shares common limitations observed in large language models, including the generation of inaccurate information ("hallucinations") and occasional repetitiveness. Because the training data are public data processed automatically, the model may have unhealthy replies. Although we try to avoid these outputs through the SFT and DPO process, it is impossible to eliminate them all. Therefore, users of our models must remain alert regarding these potential issues.

\bibliography{anthology,custom}
\bibliographystyle{acl_natbib}

\end{document}